\documentclass[10pt,a4paper]{article}

\usepackage{cogsci}

\usepackage{hyperref}
\usepackage{booktabs}
\usepackage[table,xcdraw]{xcolor}
\usepackage{makecell}
\usepackage{multirow}
\usepackage{graphicx}
\usepackage{pslatex}
\usepackage{apacite}
\usepackage{float} % Roger Levy added this and changed figure/table
\usepackage{natbib}
\usepackage{amsmath}

\title{Encoding EEG Signals to Examine Human-Like Next-Word Prediction Behaviour in Language Models}

\author{
\begin{tabular}{ccc}
\textbf{Boi Mai Quach} & \textbf{Binh T. Nguyen} & \textbf{Cathal Gurrin} \\
mai.quach3@mail.dcu.ie & ngtbinh@hcmus.edu.vn & cathal.gurrin@dcu.ie \\
School of Computing, ML-Labs, & Department of Computer Science, & School of Computing, ADAPT Centre,\\
Dublin City University, Ireland & VNUHCM – University of Science & Dublin City University, Ireland \\
\end{tabular}
\\
\\
\hspace{-0.75cm} % Adjust this value to move left or right
\begin{tabular}{ccc}
 & \textbf{Graham Healy} & \\
 & graham.healy@dcu.ie &  \\
 & School of Computing, ADAPT Centre, &\\
 & Dublin City University, Ireland &
\end{tabular}
}

\begin{document}

\maketitle

\begin{abstract}
Language models (LMs) are trained to excel at predicting the next word in the sequence given prior context, and humans also share this predictability in reading comprehension. Neuroscience research reveals that next-word predictability influences brain response, as recorded at millisecond resolution using electroencephalography (EEG). While our evidence indicates that advanced LMs achieve accuracies closely aligned with human performance at the next-word prediction task, this raises the question: Does higher prediction accuracy necessarily mean that these models adequately capture the cognitive signals associated with human reading comprehension? Here, we generate regressors for both humans and LMs based on two information measures, including top-1 prediction and surprisal, to predict event-related potential (ERP) elicited from EEG recordings which reflect different stages of cognitive processing during reading. We argue that modelling ERP patterns offers fine-grained analysis of the cognitive plausibility of various LMs during reading. Our results indicate that only surprisal potentially correlates with language-processing ERPs, especially for open-class words with high semantic content. Moreover, our findings challenge the assumption that scaling LMs with increased parameters and computational budgets will consistently lead to improved convergence with human-like linguistic processing.

\end{abstract}

\section{Introduction}
Word predictability reflects a broader cognitive process in which individuals continuously integrate context to anticipate upcoming events and test those predictions against perceptual input from the utterances they hear or read \citep{bar2007proactive}. During reading, predictability effects are hypothesized to reflect the cognitive demands associated with probabilistic inference, whereby the brain incrementally evaluates and updates possible upcoming word interpretations \citep{shain2024large}. From an information-theoretic perspective \citep{shannon1948mathematical}, prediction serves as a core function of a probabilistic, generative cognitive system that incrementally processes the words of an unfolding sentence. Linguistic units convey quantifiable information, with measures such as surprisal (the unexpectedness of a word given its prior context). Thus, surprisal serves as a useful metric for quantifying word-by-word predictability during sentence processing \citep{hale2016information}.

Research has indicated that surprisal is a reliable predictor of neural responses during reading, particularly in relation to the N400 component. \citet{michaelov2020well} found that surprisal effectively predicts variations in N400 amplitude, a neural indicator of processing difficulty during language comprehension. \citet{frank2013word} further supported these findings by analysing EEG data from participants reading identical sentences and examining four distinct ERP components. Their results highlighted that surprisal estimates significantly predict N400 amplitude, with more surprising words eliciting larger negative N400 responses.

%Give examples for using surprisal
Surprisal modelling from LMs has been widely used to explain neural responses during language comprehension, with early work showing that trigram-based surprisal correlates with the N400 in reading studies \citep{frank2015erp,willems2016prediction,armeni2019frequency}. More recent work with transformer-based LMs reports similar effects: \citet{heilbron2019tracking} showed that GPT-2’s word-by-word unpredictability aligns with N400-like responses during audiobook listening, while \citet{michaelov2024strong} found that GPT-3 surprisal best predicted N400 amplitude across models, suggesting that effects of expectancy, plausibility, and contextual semantic similarity can be explained by variations in word predictability. Although GPT-2 outperforms TransformerXL and XLNet in psycholinguistic prediction \citep{hao2020probabilistic}, evidence that larger models yield stronger predictability–processing relationships is mixed. Within the GPT family, \citet{shain2024large} challenged claims that more advanced LMs should exhibit stronger logarithmic relationships between contextual predictability and processing difficulty. They found that surprisal estimates from GPT-3 were not more ``super-logarithmic'' than those from smaller models like GPT-2, despite GPT-3's greater size and computational power.

Despite major advances in Natural Language Processing (NLP), LMs still lack an interpretable, mechanistic account aligned with how the human brain processes language. This raises debate about whether LMs capture aspects of human intelligence or simply produce outputs that mimic human thought \citep{mitchell2023debate}. Next-word predictability is a fundamental aspect of human language processing, which importantly supports the cognitive plausibility of LMs \citep{keller2010cognitively}. When it comes to thought, we should examine brain activity. During language comprehension, the human brain exhibits systematic patterns of neural activity that reflect ongoing predictive processing \citep{fitz2019language}. Therefore, rather than examining next-word prediction performance across various LMs, we investigate the relationship between next-word predictability and neural responses in natural reading contexts, especially in longer narratives.

To investigate this, we run multiple experiments. First, we calculate top-1 prediction and lexical surprisal at the word level for content and function words across three predictors: human subjects, n-gram models, and GPT-family models, using the DERCO dataset, a language resource combining EEG and next-word prediction data \citep{quach2024derco}. Next, we encode neural responses using regression-based deconvolution to estimate predictability effects on neural activity. We then compare the correlations between neural response predictions derived from top-1 prediction and surprisal estimates of language models and those obtained from human cloze responses. The purpose of this comparison is to identify which model most closely mirrors human-like predictability in reading behaviour. To provide deeper insights, these correlations will be visualised within significant time windows and across significant electrode clusters.

\section{Methodology}
\subsection{Stimuli and EEG Data Preparation}
We utilised the DERCo dataset \citep{quach2024derco}, which contains EEG recordings from 22 native English speakers while they were reading The Grimm Brothers’ Fairy Tales. Two participants (``QPF42'' and ``USQ95'') were excluded due to excessive eye movements. Additionally, word-by-word cloze probabilities were collected through a cloze procedure on Mechanical Turk crowdsourcing platform. 

EEG data were recorded using a 32-channel electrode scalp following the international 10–20 system \citep{klem1999ten}. Since the analysis used preprocessed data, the number of word-level EEG trials in the DERCo dataset's transcript was reduced due to artifact removal. All remaining words, after preprocessing, served as stimuli for encoding brain signals. The Python library IPA was used to extract parts of speech, which were then grouped into content and function words.

\subsection{Information-theoretic measures}
To investigate next-word predictability, we used two measures: top-1 prediction and surprisal. These measures serve as proxies for human and computational models' expectations and processing effort in reading comprehension, capturing different aspects of cognitive load associated with prediction.

\subsubsection{Top-1 Prediction Estimation}
\label{subsec: top1_acc_formula}
Most LMs aim to estimate a probability distribution over the vocabulary $V$, for the likely next-word $w_i \in V$ at position $i$, given the context $w_1, w_2 \ldots w_{i-1}$ containing the sequence of preceding words in text. The highest probability, also known as top-1 prediction, for the next token, is defined as follows:
\begin{equation}
    P_{w_i} = \max_{w_i \in V} P(w_i \mid w_1, w_2, \ldots, w_{i-1})
\end{equation}

\subsubsection{Surprisal Estimation}
\label{subsec: surprisal_formula}
Surprisal is a measure of the unexpectedness of a target word. \citet{hale2001probabilistic} and \citet{levy2008expectation} argued that the less expected a word is in a given context, the higher its surprisal. For example, ``Peter won the championship. Afterward, he was in seventh ...''. If readers recognise the idiom, they can guess that the missing word is ``heaven.'' Since the word is highly predictable, it has low surprisal and conveys minimal new information.

After the first t words of the sentence, $w_{1...t}$, will be processed, the identity of the upcoming word, $w_{t+1}$, is still unknown and can therefore be viewed as a random variable. The surprisal is defined as the negative log probability of the actual next word, given its preceding context:
\begin{equation}
    \textit{Surprisal } S_{w_{t+1}} = -log P(w_{t+1}|w_{1...t})
\end{equation}

\subsection{Predictors}
\subsubsection{Human Prediction}
Top-1 prediction refers to the highest percentage of participants who guessed the same next word. A top-1 prediction value of 100\% indicates that all participants guessed the same next word, whereas 0\% indicates that no participant predicted that upcoming word. This can be simply defined as the maximum cloze probability among the possible words that could appear in the upcoming position. 

Lexical surprisal, by contrast, is the cloze probability of the correct next word in the transcript. It is important to note that the cloze value of the correct next word does not necessarily equal the top-1 prediction value. These values are equal only if the word is exceptionally easy to predict, meaning that the word predicted by all participants is also the correct word in the transcript.

A major issue with cloze procedures is that zero-probability responses yield undefined surprisal values. This occurs when no participant predicts the accurate target word $w_i$ given the preceding context $w_1, w_2, \ldots, w_{i-1}$. With realistic sample sizes, words with $P(w_i \allowbreak | \allowbreak w_1, \allowbreak w_2, \allowbreak \ldots, \allowbreak w_{i-1}) < 0.001 $ are absent from responses, motivating an expanded probability distribution to include more words. Accordingly, following \citet{lowder2018lexical}, we replace zero cloze probabilities with half the smallest nonzero value before computing surprisal.

\subsubsection{N-gram models}
In this study, we trained bigram to quadgram models using the NLTK Python package \footnote{\url{https://www.nltk.org/}}. Unlike advanced language models such as transformer-based LMs \citep{amaratunga2023nlp,desai2023large}, n-gram models have a limitation in capturing very long-range dependencies. To mitigate this, we trained our n-gram models on the Fairy Tale Corpus \citep{lobo2010fairy}, a domain-aligned dataset to DERCo, comprising 453 fairy tales from Project Gutenberg \citep{klein2002fast}.

To mitigate overfitting, we excluded the five Grimm Brothers' Fairy Tales used in the DERCo dataset's transcript. All punctuation was removed, and the letters were converted to lowercase. To address data sparsity, we trained separate n-gram models with no smoothing, Laplace smoothing, and Kneser-Ney smoothing. Each model then used a fixed-sized context window of \textit{n - 1} words, locating all matching windows within the problem instance and counting the number of occurrences of each possible next token to calculate word probabilities for top-1 prediction and surprisal estimations.

\subsubsection{GPT-2 and GPT-Neo Families}
Generative Pre-trained Transformer (GPT) \citep{radford2018improving} is a transformer-based autoregressive language model that uses a multi-head ``attention'' mechanism based on an encoder-decoder architecture \citep{vaswani2017attention}. We selected GPT-2 and GPT-Neo families for investigation because they are mainly trained to predict the upcoming tokens based on probability in a left-to-right manner, similar to the cloze procedure conducted in next-word prediction tasks \citep{taylor1953cloze}. 

The transformer models take token sequences (e.g., words, phonemes, or punctuation) as input, with context window depends on the story length. If a story exceeds one context window (e.g., 1024 tokens), probability estimates for subsequent tokens are conditioned on the second half of the previous window. Predictions were performed separately for each story, with context reset between stories rather than carrying over the context from the previous one. The story's topic was included in the prediction, as it was disclosed to participants at the beginning of the experiment in the DERCo dataset. 

For words spanning multiple tokens, the word probability was calculated as the joint probability of the tokens using the chain rule. Surprisal was obtained by summing token-level surprisal values, reflecting the online experiment in which participants viewed words together with associated punctuation. In contrast, top-1 prediction focused only on correct word identification; since participants typed only the word without punctuation, the corresponding probability was averaged across the constituent tokens.

Models were implemented in PyTorch using the transformer modules from the HuggingFace Hub \footnote{https://huggingface.co/}. The examined GPT family variants differed primarily in their size, with the specific hyperparameters outlined in Table~\ref{tab:gpt_variants}.

\begin{table}[ht!]
    \centering
    \resizebox{0.36\textwidth}{!}{ % Scale table to 90% of text width
    \begin{tabular}{lcccc}
        \toprule
        \textbf{Model Name} & \textbf{$n_\text{layers}$} & \textbf{$n_{\text{head}}$} & \textbf{$d_\text{model}$} & \textbf{$n_{\text{params}}$}\\
        \midrule
        GPT-2 Small & 12 & 12 & 768 & $\sim$124M \\
        GPT-2 Medium & 24 & 16 & 1024 & $\sim$355M \\
        GPT-2 Large & 36 & 20 & 1280 & $\sim$774M \\
        
        \midrule
        GPT-Neo 125M & 12 & 12 & 768 & $\sim$125M \\
        GPT-Neo 1.3B & 24 & 16 & 2048 & $\sim$1.3B \\
        GPT-Neo 2.7B & 32 & 16 & 2560 & $\sim$2.7B \\
        
        \bottomrule
    \end{tabular}
    }
    \caption{Hyper-parameters of GPT-2, GPT-Neo families. \textbf{$n_\text{layers}$}, \textbf{$n_{\text{head}}$}, \textbf{$d_\text{model}$}, and \textbf{$n_{\text{params}}$} respectively refer to the number of layers, number of attention heads per layer, embedding size, and number of parameters.}
    \label{tab:gpt_variants}
\end{table}

\subsection{Brain Encoding Models}

%Figure 1
\begin{figure*}[!ht]
    \centering
    \includegraphics[width=.79\textwidth, height = 0.44\textwidth]{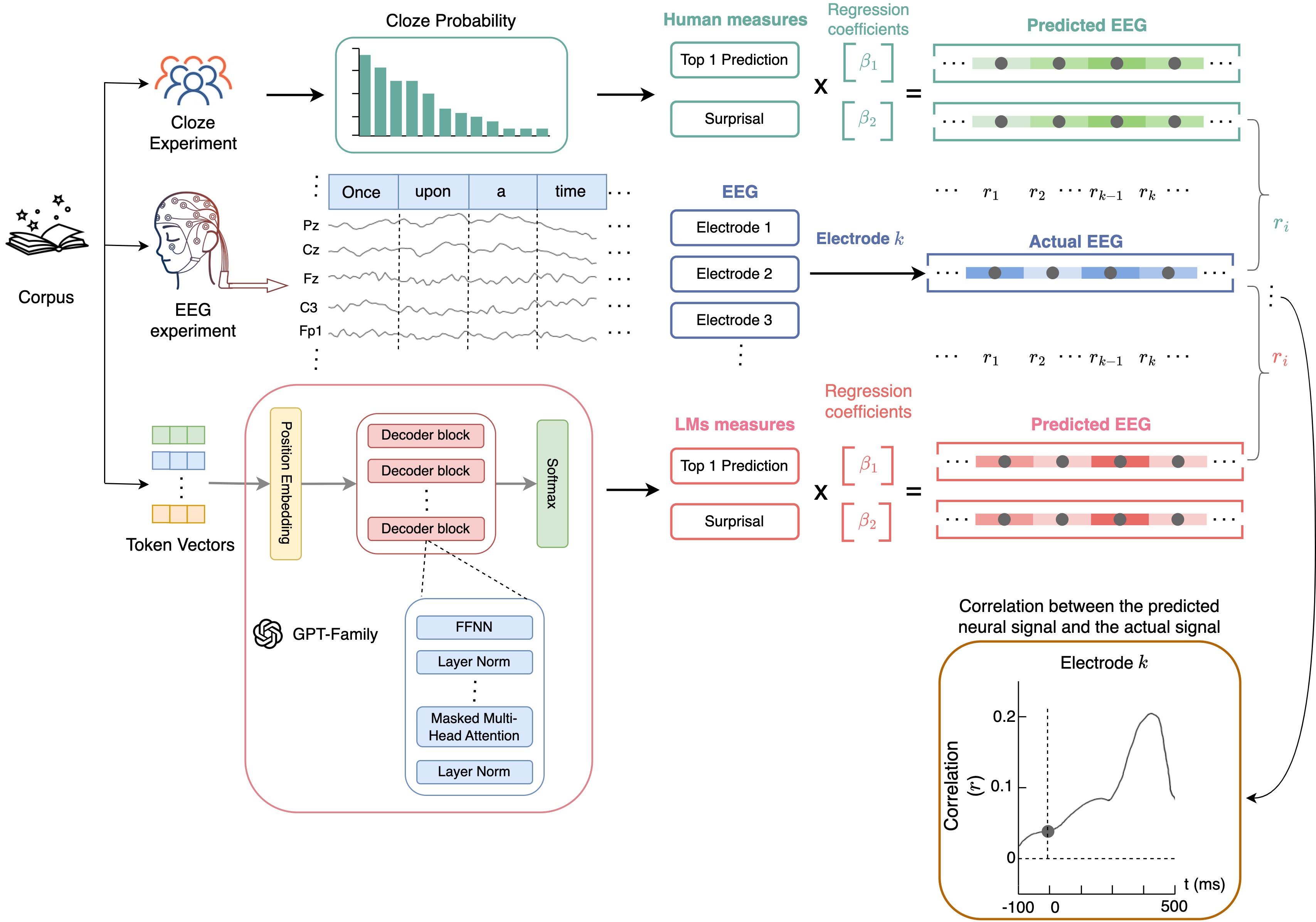}
    \caption{Brain encoding model was used to predict the neural responses to each word in the context.}
    \label{fig:brain_encoding_model}
\end{figure*}

Brain encoding models \citep{heilbron2019tracking,goldstein2022shared} entail fitting a regressor to predict neural responses at each time point based on measures such as lexical surprisal and top-1 prediction for each participant, as used in this study. However, running separate mixed effects models for each time point, as in prior studies \citep{frank2013word,hao2020probabilistic,oh2024frequency}, would require estimating an excessive number of parameters, potentially leading to model overfitting through the capture of noise rather than meaningful patterns. To reduce this, we use ridge regression, introduced by \citet{hoerl1970ridge}, which regularizes the model to control its variability and improve prediction reliability \citep{bishop2006pattern}.

Figure~\ref{fig:brain_encoding_model} illustrates the procedure for predicting EEG data using information-theoretic measures from the cloze experiment and LMs. In brief, words from the DERCo transcript were aligned with the EEG recordings from the EEG-based reading experiment (serving as the ground truth), with each word onset marked as time point 0 ms to standardise timing across trials. Two measures, top-1 prediction and surprisal, were used separately as independent variables to build regressors for predicting the EEG signal in a word-by-word, time-resolved manner. After running regressions for each measure, we estimated a series of predicted EEG amplitudes from $t_{min}$ (-100 ms) to $t_{max}$ (500 ms) for each subject per electrode. 

We quantified the similarity between the predicted and actual EEG responses using Pearson's correlation coefficient ($r$) at the word level to validate the model’s ability to capture neural representations. Using 5-fold cross-validation, each regressor was trained on 80\% of the data to learn 600 coefficients for each combination of time point and sensor, and then tested on the remaining 20\% of held-out words.

To ensure consistent regression parameters across subjects, we selected the regularisation hyperparameter $\lambda$ by fitting a single model to pooled data from all participants across a log-spaced range [$10^{-5}, 10^{-4}, …, 10^{5}$]. The optimal $\lambda$ was chosen yielding the lowest generalisation error, as estimated via 5-fold cross-validation shuffled trials, using the Scikit-learn library \citep{pedregosa2011scikit}.

\subsection{Significance Tests}
% All statistical analyses used two-tailed tests with significance levels of $\alpha = 0.05, 0.01, $ or $0.001$, depending on the desired confidence interval. Prior to univariate testing, we ensured that our data met two main assumptions: (1) normality and (2) absence of outliers. Normality was assessed using the D'Agostino-Pearson test, while outlier detection was based on the 1.5 IQR (interquartile range) method. If both assumptions were met, we applied one-sample $t$-test; otherwise, non-parametric alternatives such as the Wilcoxon signed-rank test were used to reduce the risk of misleading inferences.

In EEG research, conducting more statistical tests increases the probability of getting a false positive result due to random chance \citep{greenland2016statistical}. This issue is significantly amplified in this work, with 32 EEG sensors and 600-time points resulting in 19,200 $t$-values per subject. To address this, we implemented cluster-based permutation tests \citep{maris2007nonparametric}, using 5,000 permutations per test to identify significant time windows for the encoding models.

A mass-univariate testing approach applies one-sample $t$-tests with a ``hat'' variance adjustment to compensate implausibly small variances ($\sigma = 10^{-3}$) \citep{ridgway2012problem}. The resulting $t$-statistics were used to compute $p$-values, which were then corrected for multiple comparisons using the false discovery rate (FDR) \citep{genovese2002thresholding} for each time point and electrode to ensure statistically consistent effects.

\section{Results}
\subsection{Performance and Human Alignment in Different Language Models}
\label{subsec:correlation_performance}
Prior studies quantified neural pattern similarity between word pairs using Pearson correlation \citep{he2022neural,goldstein2022shared}, as it captures neural patterns similarity regardless of their amplitude. To strengthen our analysis, we also examined the Spearman correlation to provide more robust evidence for the observed relationships.

\begin{figure*}[!ht]
    \centering
    \includegraphics[width=0.60\textwidth, height = 0.21\textwidth]{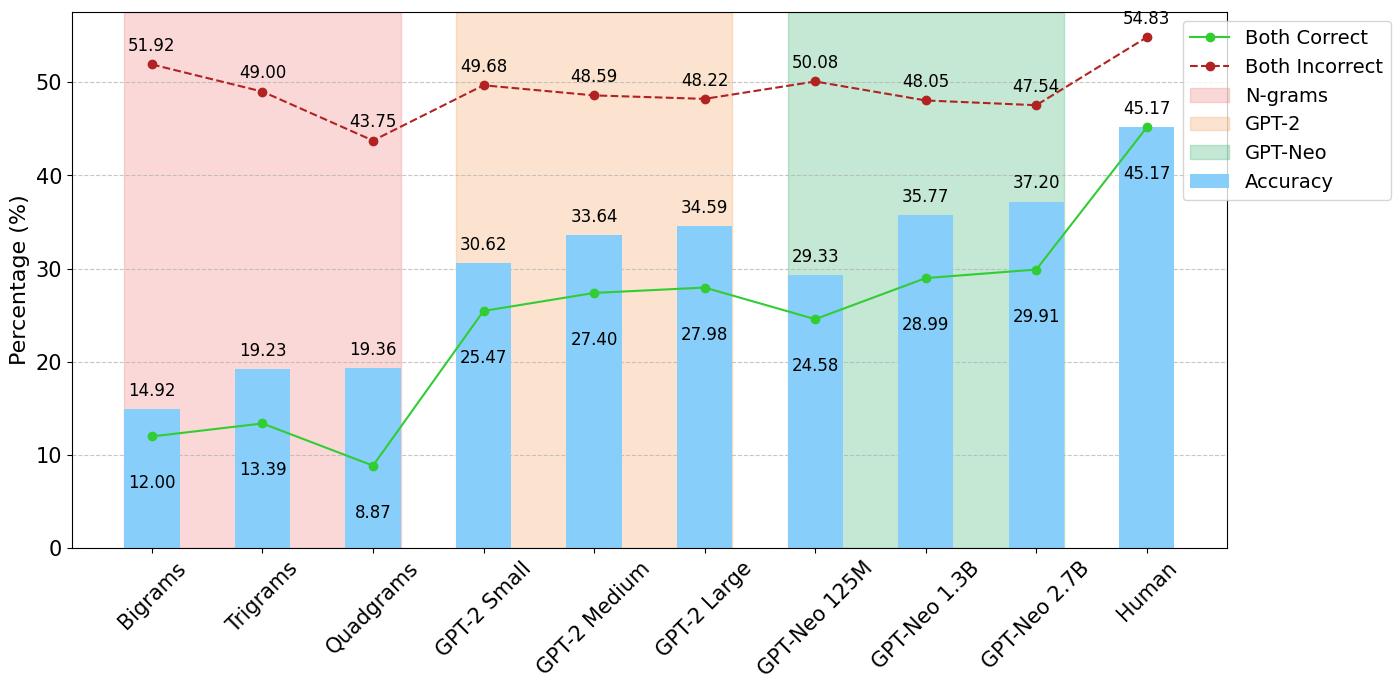}
    \caption{Performance comparison of LM families (N-grams, GPT-2, GPT-Neo) and human benchmarks in a next-word prediction task, evaluated by per-model accuracy and joint prediction percentages ``both correct'' (green) and ``both incorrect'' (red).}
    \label{fig:performance_comparison}
\end{figure*}

Table~\ref{tab:model_result_variants} presents the correlation between LMs and human predictions in terms of top-1 prediction and surprisal, along with their performance in the next-word prediction task, as measured by accuracy. Figure~\ref{fig:performance_comparison} further visualises differences in accuracy and joint accuracy between LMs and human predictions. These results yield several important observations.

\begin{table*}[!htp]
    \centering
    \scalebox{0.66}{ % Scale the table to 85% of its original size
    \begin{tabular}{lccccccccc}
        \toprule
        \textbf{Model Variants} &
        \multicolumn{2}{c}{\makecell{\textbf{Top-1 Prediction}\\\textbf{(vs. Human)}}} &
        \multicolumn{2}{c}{\makecell{\textbf{Surprisal}\\\textbf{(vs. Human)}}} &
        \multicolumn{2}{c}{\makecell{\textbf{Top-1 Prediction}\\\textbf{(vs. Human)}}} &
        \multicolumn{2}{c}{\makecell{\textbf{Surprisal}\\\textbf{(vs. Human)}}} &
        {\makecell{\textbf{Accuracy}\\\textbf{(\%)}}}
       \\
       \cmidrule(lr){2-3} \cmidrule(lr){4-5} \cmidrule(lr){6-7} \cmidrule(lr){8-9}
        & \textbf{Pearson r} & \textbf{p} & \textbf{Pearson r} & \textbf{p} & \textbf{Spearman r} & \textbf{p} & \textbf{Spearman r} & \textbf{p} \\
        
        \midrule
        \textbf{Bigrams (KN)} & 0.09 & $< p^*$ & 0.41 & $< p^*$ & 0.04 & $0.02$ & 0.41 & $< p^*$ & 14.92 \\
        \textbf{Trigrams (KN)} & 0.14 & $< p^*$ & 0.26 & $< p^*$ & 0.12 & $< p^*$ & 0.28 & $< p^*$ & 19.23 \\
        \textbf{Quadgrams (KN)} & 0.05 & $0.01$ & 0.15 & $< p^*$ & 0.02 & 0.22 & 0.15 & $< p^*$ & 19.36\\
        
        \midrule
        \textbf{GPT-2 Small} & 0.52 & $< p^*$ & 0.73 & $< p^*$ & 0.51 & $< p^*$ & 0.78 & $< p^*$ & 30.62 \\        \textbf{GPT-2 Medium} & 0.56 & $ < p^*$ & 0.75 & $< p^*$ & 0.55 & $< p^*$ & 0.80 & $< p^*$ & 33.64 \\
        \textbf{GPT-2 Large} & 0.59 & $< p^*$ & 0.77 & $< p^*$ & 0.57 & $< p^*$ & 0.82 & $< p^*$ & 34.59\\
        
        \midrule
        \textbf{GPT-Neo 125M} & 0.52 & $< p^*$ & 0.74 & $< p^*$ & 0.51 & $< p^*$ & 0.78 & $< p^*$ & 29.33\\
        \textbf{GPT-Neo 1.3B} & 0.59 & $< p^*$ & 0.77 & $< p^*$ & 0.58 & $< p^*$ & 0.82 & $< p^*$ & 35.77\\
        % \rowcolor{grey!20} % Highlight row with a light yellow color
        \textbf{GPT-Neo 2.7B} & 0.61 & $< p^*$ & 0.77 & $< p^*$ & 0.59 & $< p^*$ & 0.83 & $< p^*$ & 37.20\\

        \midrule
        \textbf{Human} & 1.00 & $0.00$ & 1.00 & $0.00$ & 1.00 & $0.00$ & 1.00 & $0.00$ & 45.17\\
        
        \bottomrule
    \end{tabular}
    }
    \caption{Correlation comparisons of various language model families against human benchmarks in a next-word prediction task. Metrics include accuracy, Pearson and Spearman correlation coefficients $r$ for top-1 prediction and surprisal. Statistically significant correlations with $p^* = 0.001$ are indicated in a two-sided test.}
    \label{tab:model_result_variants}
\end{table*}

%Figure 2

First, humans still outperformed these LMs, achieving the highest accuracy (45.17\%), highlighting the gap between LMs and human predictive capabilities. The results indicate that LMs predict next words similarly to humans in narrative contexts, with their performance gradually approaching that of humans; larger LMs are more accurate than smaller ones, but they have not surpassed human performance. 

Among n-gram models, quadgrams achieved the highest accuracy, but trigrams showed the strongest correlation with humans in top-1 predictions, while bigrams had the highest surprisal correlation. Overall, correlations between n-gram models and human predictions were generally weak and inconsistent. These results show that while historically important, n-gram models struggle to capture the complexity of human-like next-word prediction.

GPT-Neo models outperformed GPT-2, with GPT-Neo 2.7B achieving the best accuracy (37.20\%) and the highest percentage of joint correct predictions (29.91\%) among these transformer-based LMs. Both model families demonstrated strong correlations with human behaviour in top-1 predictions and surprisal metrics, with performance improving as the model size increases. Consistent increasing in both Pearson's and Spearman's $r$ correlations indicate a robust, distribution-independent linear relationship, suggesting larger models better capture human-like linguistic processing as evidenced by improved predictive accuracy and joint performance metrics.

Although advanced LMs improve significantly with scale, the mechanisms underlying their correlation with human behaviour remain unclear. \textit{Do these models truly reflect human reading processes, or do they just exhibit surface-level convergence in next-word prediction?} To investigate this, we analyse and compare results from brain encoding models, using information-theoretic measures estimated by these LMs, as detailed in the following sections.

\subsection{Neural Encoding Using Predictive Metrics from Human Cloze}
%Figure 3
\begin{figure*}[!ht]
    \centering
    \includegraphics[width=0.65\textwidth, height = 0.20\textwidth]{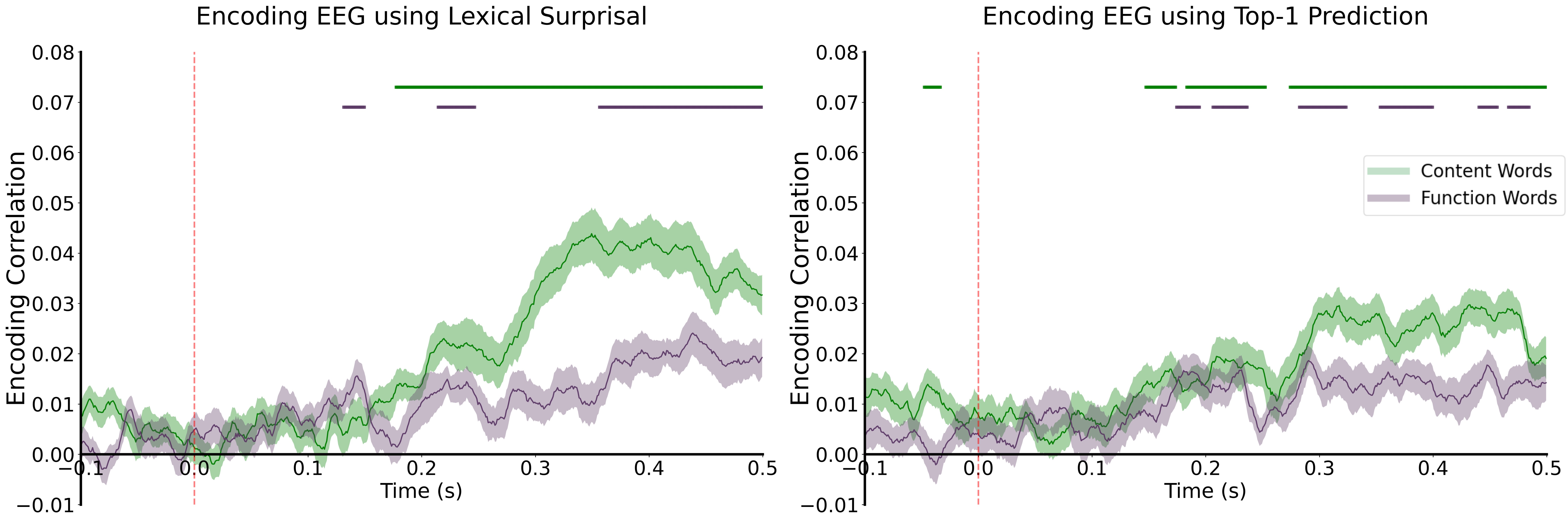}
    \caption{Correlations between human cloze–based regressors and neural responses for content and function words over time, shown for lexical surprisal (left) and top-1 prediction (right). Encoding analysis was conducted per electrode and then averaged across electrodes. Asterisks denote significant time windows ($p < 0.001$) based on cluster-based permutation tests. The shaded regions represent the between-subject standard error of the mean (SEM) of the encoding models.}
    \label{fig:human_comparison}
\end{figure*}

Figure \ref{fig:human_comparison} shows that lexical surprisal derived from human cloze probabilities is a stronger predictor of neural responses, particularly within the N400 time window, compared to the top-1 prediction measure. These results align with the well-established semantic effects on N400 amplitude \citep{frank2013word,michaelov2020well}. Additionally, encoding correlations are stronger for content words than function words, suggesting that surprisal more effectively captures neural processes associated with semantically rich lexical words \citep{munte2001differences,he2022neural}.

Therefore, we use human cloze results as the baseline for quantifying LMs' next-word predictability, lexical surprisal as the most informative metric, and mainly focus on content words to maximise analytical sensitivity.  

%Figure 6.4
\begin{figure*}[!htp]
    \centering
    \includegraphics[width=0.64\textwidth, height = 0.25\textwidth]{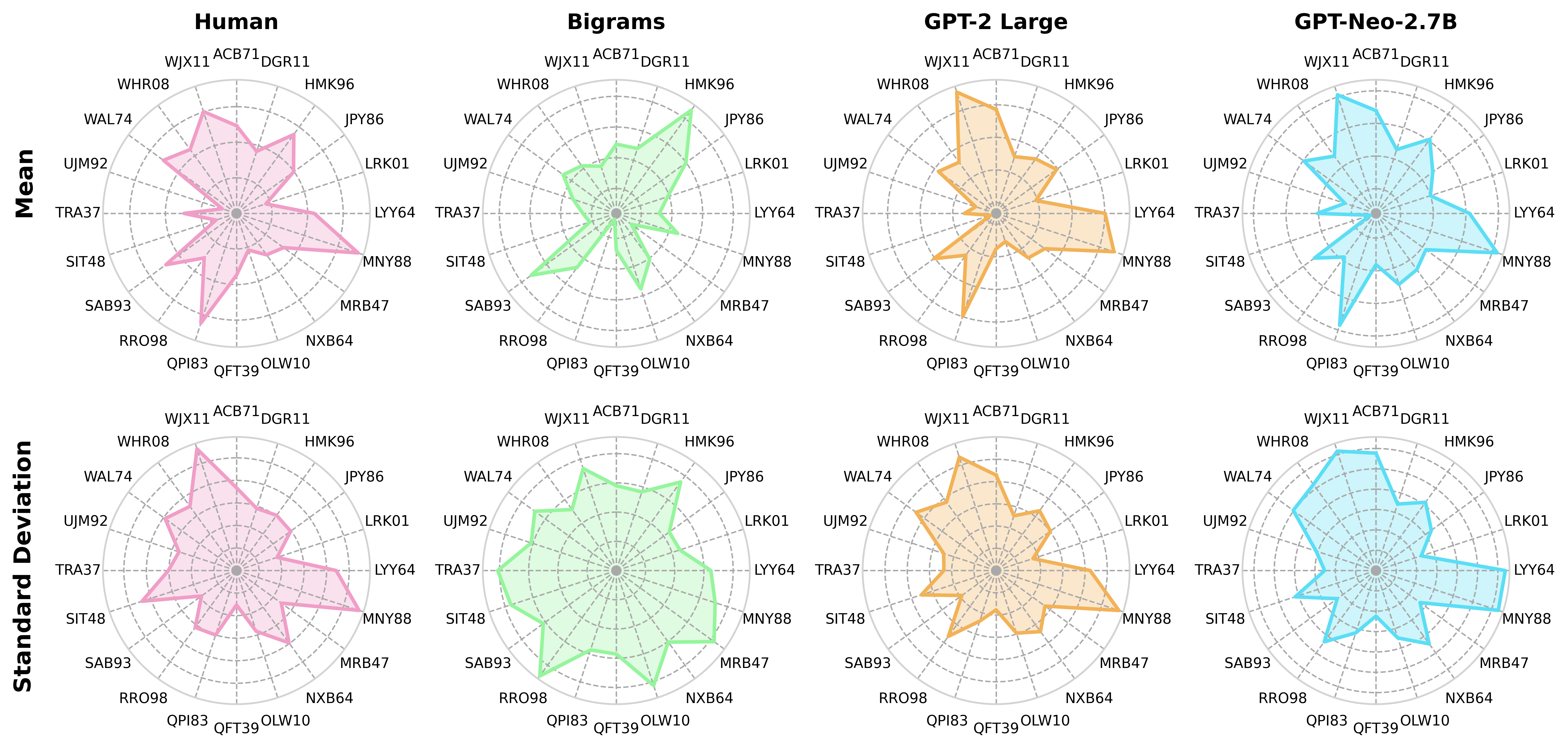}
    \caption{Radar plots of the mean (top) and standard deviation (bottom) of cross-subject surprisal correlations for content words, averaged over electrodes.}
    \label{fig:radar_representative}
\end{figure*}

%Figure 6.7
\begin{figure*}[!ht]
    \centering
    \includegraphics[width=.70\textwidth, height = .348\textwidth]{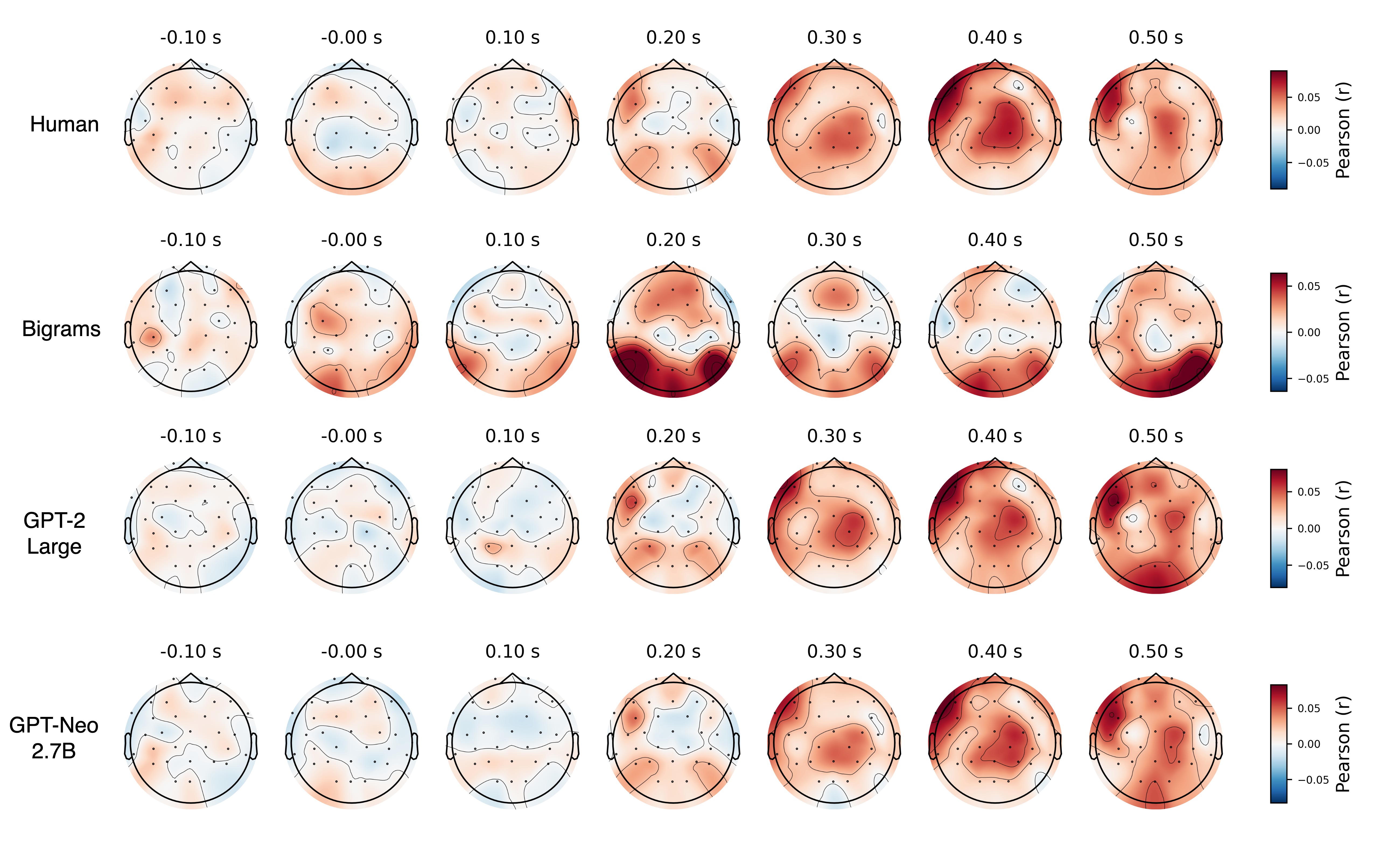}
    \caption{Full EEG Topographies of grand-averaged encoding correlations (Pearson's $r$) for lexical surprisal, computed by human predictive modelling and representative LMs over time.}
    \label{fig:topo_content_surprisal}
\end{figure*}

\subsection{Encoding Performance Comparison}
We examined encoding performance at the individual-subject level and selected Bigrams, GPT-2 Large, and GPT-Neo 2.7B as representative models for each LM family based on their top performance within their groups (see Figure~\ref{fig:performance_comparison}). As shown in Figure~\ref{fig:radar_representative}, the GPT-2 Large regression model shows the strongest correlations, with its mean and standard deviation for individual subjects closely aligning with human predictive models. Both the overall shape and the pattern of encoding correlation variances (i.e., the increase or decrease in values) are similar between GPT-2 Large and the human regressor.

\subsection{Topographic EEG analysis}
Significance levels of observed differences are reported in Figure~\ref{fig:topo_content_surprisal}. The transformer-based LMs can effectively track human neural signals using surprisal metric, particularly in predicting the content words compared to n-grams. Additionally, surprisal-based GPT-2 shows the best alignment with the encoding results from the cloze procedure. 

In the early time window (\textit{-100-100 ms}), no significant spatial pattern emerges, suggesting diffuse neural processing. However, in the \textit{200–300 ms} window, changes in correlation are observed in the centro-frontal and parieto-occipital regions, in line with the view that the P200 component reflects the processing of unexpected or affectively salient linguistic information \citep{raney1993monitoring,leuthold2015online}. In the \textit{300–500} ms range, stronger encoding correlations are observed, aligning with the N400 effect, which is associated with the ``expectedness'' level representation on the processing of upcoming words \citep{hoeks2004seeing,ye2022towards}.

\section{Discussion and Conclusion}

\label{subsec:conclusion}
This study investigated how LM predictions relate to neural responses during reading using surprisal and top-1 prediction, and their alignment with human next-word predictability. The findings indicated that surprisal-based predictors showed significant differences in neural responses for these two lexical categories, whereas top-1 prediction failed to capture this difference. Furthermore, although larger and more advanced language models typically showed a close correspondence to human productions in next-word prediction in terms of information-theoretic measures, our results demonstrated that larger model size and increased computational resources may not reliably produce more human-like language processing. Notably, surprisal-based GPT-2 Large regression substantially outperformed larger and more advanced language models in both overall and individual subject-level analyses.

Despite these insights, several limitations must be acknowledged to guide future improvements. First, our analysis focused mainly on GPT-family transformer models; however, the rapidly evolving LM landscape includes other unidirectional architectures, such as LLaMA \citep{touvron2023llama} and DeepSeekMoE \citep{dai-etal-2024-deepseekmoe}, whose distinct training objectives may impact cognitive modelling. Future work should therefore examine a broader range of transformer-based models. Second, we examined next-word predictability only via lexical groups, capturing a limited aspect of human reading behaviour. Prior work suggests that surprisal sensitivity also varies with word frequency and predictability \citep{van2019can,michaelov2020well,xia2023training,oh2024frequency}, potentially limiting generalisability across word difficulty levels. Finally, our analyses emphasised top-down semantic processing, whereas predictability estimates derived from language models reflect a combination of semantic and syntactic information \citep{qian2019neural,wilcox2021targeted,arehalli2022syntactic}; future work should incorporate syntactic contributions.

\section{Acknowledgements}
This publication has emanated from research conducted with the financial support of Science Foundation Ireland under Grant number 18/CRT/6183 and 13/RC/2106\_P2. We would like to thank the anonymous reviewers for their helpful remarks.

\section{Code Availability}
All the scripts for analysis can be found at \url{https://github.com/Tayerquach/brain_encoding_model}.

\bibliographystyle{apacite}

\setlength{\bibleftmargin}{.125in}
\setlength{\bibindent}{-\bibleftmargin}

\bibliography{CogSci}

@article{mitchell2023debate,
  title={The debate over understanding in AI’s large language models},
  author={Mitchell, Melanie and Krakauer, David C},
  journal={Proceedings of the National Academy of Sciences},
  volume={120},
  number={13},
  pages={e2215907120},
  year={2023},
  publisher={National Acad Sciences}
}

@inproceedings{keller2010cognitively,
  title={Cognitively plausible models of human language processing},
  author={Keller, Frank},
  booktitle={Proceedings of the ACL 2010 Conference Short Papers},
  pages={60--67},
  year={2010}
}

@article{fitz2019language,
  title={Language ERPs reflect learning through prediction error propagation},
  author={Fitz, Hartmut and Chang, Franklin},
  journal={Cognitive Psychology},
  volume={111},
  pages={15--52},
  year={2019},
  publisher={Elsevier}
}

@article{goldstein2022shared,
  title={{Shared computational principles for language processing in humans and deep language models}},
  author={Goldstein, Ariel and Zada, Zaid and Buchnik, Eliav and Schain, Mariano and Price, Amy and Aubrey, Bobbi and Nastase, Samuel A and Feder, Amir and Emanuel, Dotan and Cohen, Alon and others},
  journal={Nature neuroscience},
  volume={25},
  number={3},
  pages={369--380},
  year={2022},
  publisher={Nature Publishing Group US New York}
}

@article{shain2024large,
  title={Large-scale evidence for logarithmic effects of word predictability on reading time},
  author={Shain, Cory and Meister, Clara and Pimentel, Tiago and Cotterell, Ryan and Levy, Roger},
  journal={Proceedings of the National Academy of Sciences},
  volume={121},
  number={10},
  pages={e2307876121},
  year={2024},
  publisher={National Acad Sciences}
}

@article{bar2007proactive,
  title={The proactive brain: using analogies and associations to generate predictions},
  author={Bar, Moshe},
  journal={Trends in cognitive sciences},
  volume={11},
  number={7},
  pages={280--289},
  year={2007},
  publisher={Elsevier}
}

@article{shannon1948mathematical,
  title={A mathematical theory of communication},
  author={Shannon, Claude Elwood},
  journal={The Bell system technical journal},
  volume={27},
  number={3},
  pages={379--423},
  year={1948},
  publisher={Nokia Bell Labs}
}

@article{hale2016information,
  title={Information-theoretical complexity metrics},
  author={Hale, John},
  journal={Language and Linguistics Compass},
  volume={10},
  number={9},
  pages={397--412},
  year={2016},
  publisher={Wiley Online Library}
}

@inproceedings{hale2001probabilistic,
  title={A probabilistic Earley parser as a psycholinguistic model},
  author={Hale, John},
  booktitle={Second meeting of the north american chapter of the association for computational linguistics},
  year={2001}
}

@article{levy2008expectation,
  title={Expectation-based syntactic comprehension},
  author={Levy, Roger},
  journal={Cognition},
  volume={106},
  number={3},
  pages={1126--1177},
  year={2008},
  publisher={Elsevier}
}

@article{taylor1953cloze,
  title={{“Cloze procedure”: A new tool for measuring readability}},
  author={Taylor, Wilson L},
  journal={Journalism quarterly},
  volume={30},
  number={4},
  pages={415--433},
  year={1953},
  publisher={SAGE Publications Sage CA: Los Angeles, CA}
}

@inproceedings{michaelov2020well,
  title={How well does surprisal explain N400 amplitude under different experimental conditions?},
  author={Michaelov, James A and Bergen, Benjamin K},
  booktitle={Proceedings of the 24th Conference on Computational Natural Language Learning},
  pages={652--663},
  year={2020}
}

@inproceedings{frank2013word,
  title={Word surprisal predicts N400 amplitude during reading},
  author={Frank, Stefan L and Otten, Leun J and Galli, Giulia and Vigliocco, Gabriella},
  booktitle={Proceedings of the 51st Annual Meeting of the Association for Computational Linguistics (Volume 2: Short Papers)},
  pages={878--883},
  year={2013}
}

@inproceedings{hao2020probabilistic,
  title={Probabilistic Predictions of People Perusing: Evaluating Metrics of Language Model Performance for Psycholinguistic Modeling},
  author={Hao, Yiding and Mendelsohn, Simon and Sterneck, Rachel and Martinez, Randi and Frank, Robert},
  booktitle={Proceedings of the Workshop on Cognitive Modeling and Computational Linguistics},
  pages={75--86},
  year={2020}
}

@article{frank2015erp,
  title={The ERP response to the amount of information conveyed by words in sentences},
  author={Frank, Stefan L and Otten, Leun J and Galli, Giulia and Vigliocco, Gabriella},
  journal={Brain and language},
  volume={140},
  pages={1--11},
  year={2015},
  publisher={Elsevier}
}

@article{willems2016prediction,
  title={Prediction during natural language comprehension},
  author={Willems, Roel M and Frank, Stefan L and Nijhof, Annabel D and Hagoort, Peter and Van den Bosch, Antal},
  journal={Cerebral cortex},
  volume={26},
  number={6},
  pages={2506--2516},
  year={2016},
  publisher={Oxford University Press}
}

@article{armeni2019frequency,
  title={Frequency-specific brain dynamics related to prediction during language comprehension},
  author={Armeni, Kristijan and Willems, Roel M and Van den Bosch, Antal and Schoffelen, Jan-Mathijs},
  journal={NeuroImage},
  volume={198},
  pages={283--295},
  year={2019},
  publisher={Elsevier}
}

@inproceedings{heilbron2019tracking,
  title={Tracking naturalistic linguistic predictions with deep neural language models},
  author={Heilbron, Micha and Ehinger, Benedikt and Hagoort, Peter and De Lange, Floris P},
  booktitle={2019 Conference on Cognitive Computational Neuroscience (CCN 2019)},
  pages={424--427},
  year={2019}
}

@article{michaelov2024strong,
  title={Strong Prediction: Language model surprisal explains multiple N400 effects},
  author={Michaelov, James A and Bardolph, Megan D and Van Petten, Cyma K and Bergen, Benjamin K and Coulson, Seana},
  journal={Neurobiology of language},
  volume={5},
  number={1},
  pages={107--135},
  year={2024},
  publisher={MIT Press One Broadway, 12th Floor, Cambridge, Massachusetts 02142, USA~…}
}

@article{quach2024derco,
  title={DERCo: A Dataset for Human Behaviour in Reading Comprehension Using EEG},
  author={Quach, Boi Mai and Gurrin, Cathal and Healy, Graham},
  journal={Scientific Data},
  volume={11},
  number={1},
  pages={1104},
  year={2024},
  publisher={Nature Publishing Group UK London}
}

@incollection{amaratunga2023nlp,
  title={NLP Through the Ages},
  author={Amaratunga, Thimira},
  booktitle={Understanding Large Language Models: Learning Their Underlying Concepts and Technologies},
  pages={9--54},
  year={2023},
  publisher={Springer}
}

@article{desai2023large,
  title={Large Language Models: A Comprehensive Exploration of Modern AI's Potential and Pitfalls},
  author={Desai, Bhavin and Patil, Kapil and Patil, Asit and Mehta, Ishita},
  journal={Journal of Innovative Technologies},
  volume={6},
  number={1},
  year={2023}
}

@inproceedings{lobo2010fairy,
  title={Fairy Tale Corpus Organization Using Latent Semantic Mapping and an Item-to-item Top-n Recommendation Algorithm},
  author={Lobo, Paula Vaz and de Matos, David Martins},
  booktitle={Proceedings of the Seventh International Conference on Language Resources and Evaluation (LREC'10)},
  year={2010}
}

@article{klein2002fast,
  title={Fast exact inference with a factored model for natural language parsing},
  author={Klein, Dan and Manning, Christopher D},
  journal={Advances in neural information processing systems},
  volume={15},
  year={2002}
}

@article{lowder2018lexical,
  title={Lexical predictability during natural reading: Effects of surprisal and entropy reduction},
  author={Lowder, Matthew W and Choi, Wonil and Ferreira, Fernanda and Henderson, John M},
  journal={Cognitive science},
  volume={42},
  pages={1166--1183},
  year={2018},
  publisher={Wiley Online Library}
}

@inproceedings{oh2024frequency,
  title={Frequency Explains the Inverse Correlation of Large Language Models’ Size, Training Data Amount, and Surprisal’s Fit to Reading Times},
  author={Oh, Byung-Doh and Yue, Shisen and Schuler, William},
  booktitle={Proceedings of the 18th Conference of the European Chapter of the Association for Computational Linguistics (Volume 1: Long Papers)},
  pages={2644--2663},
  year={2024}
}

@book{bishop2006pattern,
  title={Pattern recognition and machine learning},
  author={Bishop, Christopher M and Nasrabadi, Nasser M},
  volume={4},
  number={4},
  year={2006},
  publisher={Springer}
}

@article{hoerl1970ridge,
  title={Ridge regression: applications to nonorthogonal problems},
  author={Hoerl, Arthur E and Kennard, Robert W},
  journal={Technometrics},
  volume={12},
  number={1},
  pages={69--82},
  year={1970},
  publisher={Taylor \& Francis}
}

@article{pedregosa2011scikit,
  title={Scikit-learn: Machine learning in Python},
  author={Pedregosa, Fabian and Varoquaux, Ga{\"e}l and Gramfort, Alexandre and Michel, Vincent and Thirion, Bertrand and Grisel, Olivier and Blondel, Mathieu and Prettenhofer, Peter and Weiss, Ron and Dubourg, Vincent and others},
  journal={the Journal of machine Learning research},
  volume={12},
  pages={2825--2830},
  year={2011},
  publisher={JMLR. org}
}

@article{greenland2016statistical,
  title={Statistical tests, P values, confidence intervals, and power: a guide to misinterpretations},
  author={Greenland, Sander and Senn, Stephen J and Rothman, Kenneth J and Carlin, John B and Poole, Charles and Goodman, Steven N and Altman, Douglas G},
  journal={European journal of epidemiology},
  volume={31},
  number={4},
  pages={337--350},
  year={2016},
  publisher={Springer}
}

@article{maris2007nonparametric,
  title={Nonparametric statistical testing of EEG-and MEG-data},
  author={Maris, Eric and Oostenveld, Robert},
  journal={Journal of neuroscience methods},
  volume={164},
  number={1},
  pages={177--190},
  year={2007},
  publisher={Elsevier}
}

@article{he2022neural,
  title={Neural correlates of word representation vectors in natural language processing models: Evidence from representational similarity analysis of event-related brain potentials},
  author={He, Taiqi and Boudewyn, Megan A and Kiat, John E and Sagae, Kenji and Luck, Steven J},
  journal={Psychophysiology},
  volume={59},
  number={3},
  pages={e13976},
  year={2022},
  publisher={Wiley Online Library}
}

@article{touvron2023llama,
  title={Llama: Open and efficient foundation language models},
  author={Touvron, Hugo and Lavril, Thibaut and Izacard, Gautier and Martinet, Xavier and Lachaux, Marie-Anne and Lacroix, Timoth{\'e}e and Rozi{\`e}re, Baptiste and Goyal, Naman and Hambro, Eric and Azhar, Faisal and others},
  journal={arXiv preprint arXiv:2302.13971},
  year={2023}
}

@inproceedings{van2019can,
  title={Can Entropy Explain Successor Surprisal Effects in Reading?},
  author={van Schijndel, Marten and Linzen, Tal},
  booktitle={Proceedings of the Society for Computation in Linguistics (SCiL) 2019},
  pages={1--7},
  year={2019}
}

@inproceedings{xia2023training,
  title={Training Trajectories of Language Models Across Scales},
  author={Xia, Mengzhou and Artetxe, Mikel and Zhou, Chunting and Lin, Xi Victoria and Pasunuru, Ramakanth and Chen, Danqi and Zettlemoyer, Luke and Stoyanov, Veselin},
  booktitle={Proceedings of the 61st Annual Meeting of the Association for Computational Linguistics (Volume 1: Long Papers)},
  pages={13711--13738},
  year={2023}
}

@inproceedings{qian2019neural,
  title={Neural language models as psycholinguistic subjects: representations of syntactic state},
  author={Qian, Peng and Levy, Roger P},
  year={2019},
  organization={Association for Computational Linguistics}
}

@inproceedings{arehalli2022syntactic,
  title={Syntactic Surprisal From Neural Models Predicts, But Underestimates, Human Processing Difficulty From Syntactic Ambiguities},
  author={Arehalli, Suhas and Dillon, Brian W and Linzen, Tal},
  booktitle={Proceedings of the 26th Conference on Computational Natural Language Learning (CoNLL)},
  pages={301--313},
  year={2022}
}

@inproceedings{wilcox2021targeted,
  title={A Targeted Assessment of Incremental Processing in Neural Language Models and Humans},
  author={Wilcox, Ethan and Vani, Pranali and Levy, Roger},
  booktitle={Proceedings of the 59th Annual Meeting of the Association for Computational Linguistics and the 11th International Joint Conference on Natural Language Processing (Volume 1: Long Papers)},
  pages={939--952},
  year={2021}
}

@article{klem1999ten,
  title={{The ten-twenty electrode system of the international federation. The international federation of clinical neurophysiology}},
  author={Klem, George H},
  journal={Electroencephalogr. Clin. Neurophysiol. Suppl.},
  volume={52},
  pages={3--6},
  year={1999}
}

@article{radford2018improving,
  title={Improving language understanding by generative pre-training},
  author={Radford, Alec},
  year={2018}
}

@article{vaswani2017attention,
  title={Attention is all you need},
  author={Vaswani, A},
  journal={Advances in Neural Information Processing Systems},
  year={2017}
}

@article{genovese2002thresholding,
  title={Thresholding of statistical maps in functional neuroimaging using the false discovery rate},
  author={Genovese, Christopher R and Lazar, Nicole A and Nichols, Thomas},
  journal={Neuroimage},
  volume={15},
  number={4},
  pages={870--878},
  year={2002},
  publisher={Elsevier}
}

@article{ridgway2012problem,
  title={The problem of low variance voxels in statistical parametric mapping; a new hat avoids a ‘haircut’},
  author={Ridgway, Gerard R and Litvak, Vladimir and Flandin, Guillaume and Friston, Karl J and Penny, Will D},
  journal={Neuroimage},
  volume={59},
  number={3},
  pages={2131--2141},
  year={2012},
  publisher={Elsevier}
}

@article{munte2001differences,
  title={Differences in brain potentials to open and closed class words: Class and frequency effects},
  author={M{\"u}nte, Thomas F and Wieringa, Bernardina M and Weyerts, Helga and Szentkuti, Andras and Matzke, Mike and Johannes, S{\"o}nke},
  journal={Neuropsychologia},
  volume={39},
  number={1},
  pages={91--102},
  year={2001},
  publisher={Elsevier}
}

@inproceedings{ye2022towards,
  title={Towards a better understanding of human reading comprehension with brain signals},
  author={Ye, Ziyi and Xie, Xiaohui and Liu, Yiqun and Wang, Zhihong and Chen, Xuesong and Zhang, Min and Ma, Shaoping},
  booktitle={Proceedings of the ACM Web Conference 2022},
  pages={380--391},
  year={2022}
}

@article{leuthold2015online,
  title={Online processing of moral transgressions: ERP evidence for spontaneous evaluation},
  author={Leuthold, Hartmut and Kunkel, Angelika and Mackenzie, Ian G and Filik, Ruth},
  journal={Social cognitive and affective neuroscience},
  volume={10},
  number={8},
  pages={1021--1029},
  year={2015},
  publisher={Oxford University Press}
}

@article{hoeks2004seeing,
  title={Seeing words in context: the interaction of lexical and sentence level information during reading},
  author={Hoeks, John CJ and Stowe, Laurie A and Doedens, Gina},
  journal={Cognitive brain research},
  volume={19},
  number={1},
  pages={59--73},
  year={2004},
  publisher={Elsevier}
}

@article{raney1993monitoring,
  title={Monitoring changes in cognitive load during reading: an event-related brain potential and reaction time analysis.},
  author={Raney, Gary E},
  journal={Journal of Experimental Psychology: Learning, Memory, and Cognition},
  volume={19},
  number={1},
  pages={51},
  year={1993},
  publisher={American Psychological Association}
}

@inproceedings{dai-etal-2024-deepseekmoe,
    title = "{D}eep{S}eek{M}o{E}: Towards Ultimate Expert Specialization in Mixture-of-Experts Language Models",
    author = "Dai, Damai  and
      Deng, Chengqi  and
      Zhao, Chenggang  and
      Xu, R.x.  and
      Gao, Huazuo  and
      Chen, Deli  and
      Li, Jiashi  and
      Zeng, Wangding  and
      Yu, Xingkai  and
      Wu, Y.  and
      Xie, Zhenda  and
      Li, Y.k.  and
      Huang, Panpan  and
      Luo, Fuli  and
      Ruan, Chong  and
      Sui, Zhifang  and
      Liang, Wenfeng",
    editor = "Ku, Lun-Wei  and
      Martins, Andre  and
      Srikumar, Vivek",
    booktitle = "Proceedings of the 62nd Annual Meeting of the Association for Computational Linguistics (Volume 1: Long Papers)",
    month = aug,
    year = "2024",
    address = "Bangkok, Thailand",
    publisher = "Association for Computational Linguistics",
    url = "https://aclanthology.org/2024.acl-long.70/",
    doi = "10.18653/v1/2024.acl-long.70",
    pages = "1280--1297"
}

\end{document}